\documentclass[letterpaper, 10 pt, conference]{ieeeconf}  
\IEEEoverridecommandlockouts                              
\usepackage{epsfig} 
\usepackage{times} 
\usepackage{hyperref}

\usepackage{amsmath} 
\usepackage{amssymb}  
\usepackage{mathrsfs}
\usepackage{algorithm,algorithmic}

\title{\Large \bf
Bipedal Hopping: Reduced-order Model Embedding via Optimization-based Control
}
    \author{Xiaobin Xiong and Aaron D. Ames
    \thanks{*This work is supported by NSF grant NRI-1526519.}
\thanks{The authors are with the Department of Mechanical and Civil Engineering, California Institute of Technology, Pasadena, CA 91125
        {\tt\small xxiong@caltech.edu}, {\tt\small ames@caltech.edu}}%
 }
\begin{document}
\maketitle
\thispagestyle{empty}
\pagestyle{empty}

\begin{abstract}
This paper presents the design and validation of controlling hopping on the 3D bipedal robot Cassie. A spring-mass model is identified from the kinematics and compliance of the robot. The spring stiffness and damping are encapsulated by the leg length, thus actuating the leg length can create and control hopping behaviors. Trajectory optimization via direct collocation is performed on the spring-mass model to plan jumping and landing motions. The leg length trajectories are utilized as desired outputs to synthesize a control Lyapunov function based quadratic program (CLF-QP). Centroidal angular momentum, taking as an addition output in the CLF-QP, is also stabilized in the jumping phase to prevent whole body rotation in the underactuated flight phase. The solution to the CLF-QP is a nonlinear feedback control law that achieves dynamic jumping behaviors on bipedal robots with compliance. The framework presented in this paper is verified experimentally on the bipedal robot Cassie.
\end{abstract}

\section{INTRODUCTION}
Reduced-order models such as the canonical Spring Loaded Inverted Pendulum (SLIP) have been widely applied for controlling walking \cite{rummel2010stable} \cite{XiongAG17} \cite{hereid2014embedding}, running \cite{raibert1986legged} and hopping  \cite{poulakakis2009spring} of legged robots. One important benefit of using low order dynamical systems for control is that it renders the gait and motion generation problems for legged robots computationally tractable. However, reduced-order models are often directly implemented on the full-order model of the robot, e.g., through inverse kinematics \cite{kajita2003biped} or inverse dynamics \cite{pratt2012capturability}, without a faithful connection to the structure and morphology of the robot.

In this paper, we present an approach to identifying the spring-mass model for bipedal robots with mechanical compliance, and synthesizing nonlinear controllers by embedding the spring-mass model into the full-order dynamics. Specifically, the spring in the spring-mass model comes from viewing each leg as a deformable prismatic spring as motivated by the mechanical design of robots with compliance \cite{hubicki2016atrias}. We borrow the idea of end-effector stiffness from manipulation community \cite{siciliano2008springer}, and formally derive the stiffness/damping of the leg spring from the compliant components in the leg as functions of robot configurations. This facilitates the spring-mass model being virtually actuated by changing robot configurations, i.e., by changing the leg length on the spring-mass model. Trajectory optimization can thus be utilized to create hopping behaviors on the spring-mass model.

The planned leg length trajectory from the spring-mass model encodes the underactuation of the leg compliance of the robot, and can therefore be used to synthesize controllers that achieve this behavior on the full-order model. Roughly speaking, the springs on the physical robot are expected to behave similarly to the spring in the spring-mass model when the robot tracks the leg length trajectory. This motivates defining the leg length trajectory as a desired output on each leg and thus formulate a control Lyapunov function based quadratic program (CLF-QP)  \cite{ames2013towards} \cite{ames2014rapidly} for output stabilization. The end result is a nonlinear optimization-based controller that represents the reduced-order dynamics in the full-order model of the robot.
\begin{figure}[t]
      \centering
      \includegraphics[height= 2.4in]{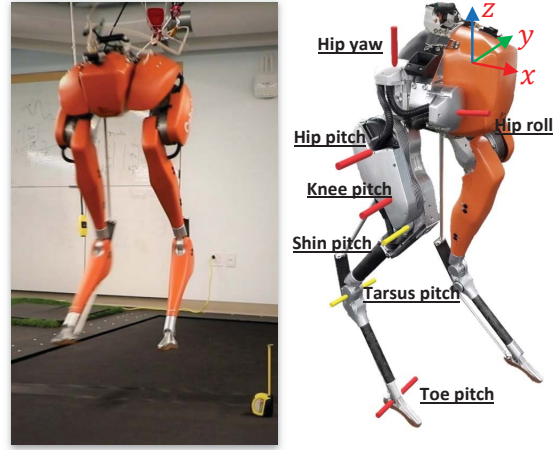}
      \caption{Hopping on Cassie \cite{Supplementary} (left) and its coordinate system (right).}
      \label{Cassie}
\end{figure}

The QP formulation for hopping is inspired by the approach for walking in \cite{hereid2014embedding}, wherein the SLIP dynamics is embedded onto the center of mass (COM) dynamics of the full robot via an equality constraint in the QP. The difference of our approach is that the hopping dynamics is embedded by taking the leg length trajectory as a desired output, which becomes an inequality CLF constraint in the QP, rendering a more feasible QP formulation. Additionally, the hopping motion naturally requires a consideration on momentum regulation due to the conservation law on centroidal angular momentum \cite{orin2008centroidal} in flight phase. This is done by including the angular momentum as an output to stabilize in the CLF-QP.

The proposed approach is successfully implemented on the 3D underactuated bipedal robot Cassie (see Fig. \ref{Cassie}) in both simulation and experiment \cite{Supplementary}, achieving the hopping on Cassie with ground clearance of $\sim7$ inches and air-time of $\sim0.423s$. The ground reaction force and toe-off timing of hopping motions on the robot match closely with these of the spring-mass model. This further indicates a faithful construction and embedding of the reduced-order model onto the full order model of the robot.

%
\section{Robot Model}
\label{sec:Modeling}
The Cassie-series robot from Agility Robotics \cite{AG} is a full 3D bipedal robot that is designed to be agile and robust. Like its predecessor ATRIAS \cite{hubicki2016atrias}, Cassie is designed with concentrated mass at its pelvis and lightweight legs with leaf springs and closed kinematic chains. The mechanical design, thus, embodies the SLIP model \cite{rummel2010stable}. From the perspective of model-based control, the compliant closed chain on each leg can, however, create additional complexities. Therefore, rigid model \cite{hereid2014embedding} or overly simplified model \cite{hubicki2016atrias} is oftentimes applied. Here, we present the full body dynamics model with justifiable simplifications.

\begin{figure}[t]
      \centering
      \includegraphics[width=3.1in]{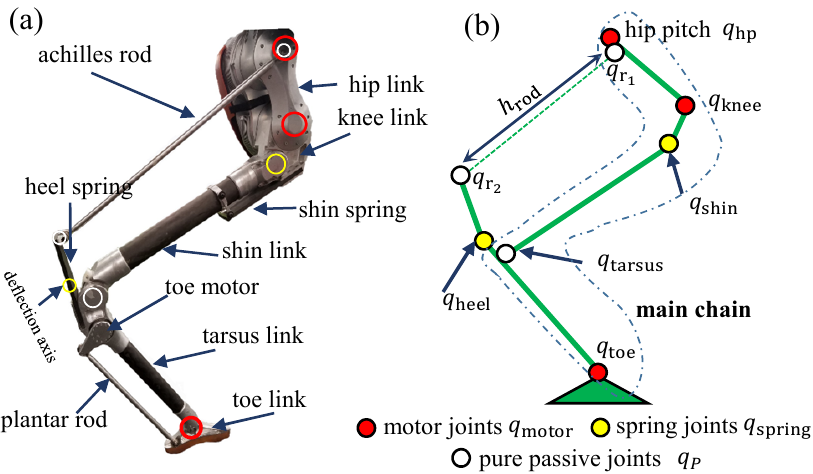}
      \caption{(a,b) Cassie's leg and its model.}
      \label{CassieLeg}
\end{figure}

As shown in Fig. \ref{Cassie}, Cassie has five motor joints (with the axis of rotation shown in red) on each leg, three of which locate at the hip and the other two are the knee and toe pitch. Fig. \ref{CassieLeg}(a) and (b) provide a close look at the leg kinematics and the abstract model, respectively. We model the shin and heel springs as torsion springs at the corresponding deflection axes. Therefore the spring torques are:
 \begin{equation}
\tau_{\textrm{shin/heel}} = k_{\textrm{shin/heel}} q_{\textrm{shin/heel}}  + d_{\textrm{shin/heel}} \dot{q}_{\textrm{shin/heel}},
 \end{equation}
where $k_{\textrm{shin/heel}},d_{\textrm{shin/heel}}$ are the stiffness and damping, provided by the manufacturer \cite{AG}. Since the achilles rod is very lightweight, we ignore the achilles rod and replace it by setting a holonomic constraint $h_{\textrm{rod}}$ on the distance between the connectors (one locates on the inner side of hip joint, the other locates at the end of the heel spring). The plantar rod is also removed and the actuation is applied to the toe pitch directly thanks to the parallel linkage design. These two simplifications removed unnecessary passive joints and associated configuration variables. As a consequence, the configuration of the leg can be described only by five motor joints, two spring joints and a passive tarsus joint. The total number of degrees of freedom of the floating base model is then $n =8 \times2 + 6 =22$. The dynamics can be derived from the Euler-Lagrange equation with holonomic constraints as:
\begin{eqnarray}
&& M(q)\ddot{q} + H(q,\dot{q}) = Bu+J_s^T \tau_s + J_{h,v}^T F_{h,v},  \label{eom} \\
&& J_{h,v}(q) \ddot{q} + \dot{J}_{h,v}(q) \dot{q} = 0, \label{hol}
\end{eqnarray}
where $q\in \mathbb{R}^{n}$, $M(q)$ is the mass matrix, $H(q,\dot{q})$ is the Coriolis, centrifugal and gravitational term, $B$ and $u\in \mathbb{R}^{10}$ are the actuation matrix and the motor torque vector, $\tau_s$ and $J_s$ are the spring joint torque vector and the corresponding Jacobian, and $F_{h,v}\in \mathbb{R}^{n_{h,v}}$ and $J_{h,v}$ are the holonomic force vector and the corresponding Jacobian respectively. The subscript $v$ is used to indicate different domains which have different numbers of holonomic constraints. For instance, when the robot has no contact with the ground, $n_{h,v} = 2$ as there are two holonomic constraints on $h_{\textrm{rod}}$. In case when the feet contact the ground, five additional holonomic constraints are introduced on each foot, hence $n_{h,v} = 12$.
\section{Spring-Mass Model}\label{SLIPmodel}
In this section, we derive our spring-mass model from the kinematics of the robot. The compliant components on the leg are characterized as a prismatic spring on the leg in the spring-mass model. It is expected that the stiffness of the leg spring changes with different robot configurations, thus we explicitly derive the leg stiffness $K_{\textrm{leg}}$ as a function of joint angles, the analogy of which is the end-effector stiffness for robotic manipulators \cite{siciliano2008springer}. With an eye towards the motion planning for the spring-mass model, $K_{\textrm{leg}}$ is approximated by a polynomial function of leg length $L$. Lastly, we present the trajectory optimization via direct collocation for the spring-mass model.
\subsection{Leg Stiffness and Leg Length}
The leg stiffness $K_{\textrm{leg}}$ is the resistance of the leg to external forces. The complementary concept is called leg compliance $C_{\textrm{leg}} = K_{\textrm{leg}}^{-1}$. When the leg is under external load at the foot, the leg deforms due to compliant elements in the leg. Assuming that we only consider the transitional deformations, the external force can be calculated by,
\begin{equation}
F_{\textrm{ext}} = K_{\textrm{leg}}\delta,
\end{equation}
where $F_{\textrm{ext}} \in \mathbb{R}^3, K_{\textrm{leg}}\in \mathbb{R}^{3 \times 3}$ and $\delta\in \mathbb{R}^3$. Under the assumption that the deformation is small and only happens at the joints, the leg deformation $\delta$ can be mapped from joint deformations $\Delta q$ by the foot Jacobian as,
\begin{equation}
\delta = J \Delta q,
\end{equation}
where $J \in \mathbb{R}^{3\times n}$ and $\Delta q \in \mathbb{R}^{n}$. Let $\tau$ denotes the moments at the joints caused by the external load, thus $\tau = J^T F_{\textrm{ext}}$. If the stiffness at each joint is $k_i$ with $i = 1, ... ,n$, then the joint stiffness matrix is defined as $K_J = \textrm{diag}(k_1, ... ,k_n)$, and $\tau = K_J\Delta q$. The joint stiffness matrix $K_J $ and leg stiffness $K_{\textrm{leg}}$ are hence related by the joint moments,
\begin{equation}
K_J\Delta q = \tau = J^T F_{\textrm{ext}} =J^T K_{\textrm{leg}}\delta = J^T K_{\textrm{leg}}J \Delta q.
\end{equation}
Then the leg stiffness can be calculated from the joint stiffness matrix by,
\begin{equation}
K_{\textrm{leg}}(q) = (J(q)K_J^{-1}J(q)^T)^{-1}.
 \end{equation}
This indicates the leg stiffness is a function of the Jacobian and thus a function of the configuration $q$. The leg damping $D_\textrm{leg}(q)$ is derived in the same way by dropping the assumption on small deformation at joints.

Now we apply the calculation of leg stiffness and damping on Cassie. Note that the main difference is the closed kinematic chain inside the leg with pure passive joints and compliant joints. The pure passive joints have no contribution towards leg stiffness, so we need to derive the forward kinematics from active joints, i.e. the spring joints and motor joints. As there are two chains towards the toe, the velocity of the toe relative to the hip can be calculated as,
\begin{equation}
v_{\textrm{Toe} \leftarrow \textrm{Hip}} = J_1(q_1) \dot{q_1} = J_2(q_2) \dot{q_2} ,\label{twoChain}
\end{equation}
where $q_1 = [ q_{\textrm{hp}}; q_{\textrm{knee}}; q_{\textrm{shin}}; q_{\textrm{tarsus}}; q_{\textrm{toe}}]$ and $q_2 =  [q_{\textrm{r}_2}; q_{\textrm{r}_1}; q_{\textrm{heel}}; q_{\textrm{toe}}]$ (see Fig. \ref{CassieLeg}). Eq. \eqref{twoChain} can be rewritten as,
\begin{equation}
\underset{E}{\underbrace{{\begin{bmatrix} J_1(q_1) & -J_2(q_2)\end{bmatrix}}}}
\begin{bmatrix}  \dot{q_1} \\  \dot{q_2}  \end{bmatrix} = 0.
\end{equation}
Then we can rearrange the matrix $E$ and group $q_1,q_2$ into active joints $q_{A}= \{q_{\textrm{spring}}, q_{\textrm{motor}} \}$ and pure passive joints $q_P$. Eq. \eqref{twoChain} becomes,
\begin{equation}
\begin{bmatrix} E_A(q) & E_P(q) \end{bmatrix}  \begin{bmatrix}
 \dot{q}_A \\  \dot{q}_P
\end{bmatrix} = 0.
\end{equation}
Then the passive joint velocity  $\dot{q}_P = -E_P^{-1}E_A\dot{q}_A$. As there is only one passive joint, the tarsus, on the main kinematic chain, we can find $\dot{q}_{\textrm{tarsus}} =  -E_P^{-1}E_A\dot{q}_A$, where $q_A = \{q_{\textrm{knee}}, q_{\textrm{shin}}, q_{\textrm{heel}}, q_{\textrm{toe}} \}$. Then the forward kinematics and the leg stiffness can be expressed in terms of $q_A$,
\begin{eqnarray}
v_{\textrm{Toe} \leftarrow \textrm{Hip}}  &=&  J_A(q_A) \dot{q}_A,\\
 K_{\textrm{leg}}(q_A)  &=& (J_A(q_A)  K_A^{-1}J_A(q_A) ^T)^{-1},
\end{eqnarray}
where $K_A = \textrm{diag}(\infty , k_{\textrm{shin}},k_{\textrm{heel}},\infty )$ as we assume the motor joints being rigidly controlled to fixed positions. Assuming the spring joints have small deflections under normal load, $K_{\textrm{leg}}(q_A)$ can be approximated as $K_{\textrm{leg}}(q_{\textrm{knee}}, q_{\textrm{shin}}=0, q_{\textrm{heel}}=0, q_{\textrm{toe}})$. $q_{\textrm{toe}}$ has trivial contribution in terms of $J_A$ and $K_{\textrm{leg}}$. Thus $K_{\textrm{leg}}(q_A)\approx K_{\textrm{leg}}(q_{\textrm{knee}})$.

\begin{figure}[t]
      \centering
      \includegraphics[width=3.3in]{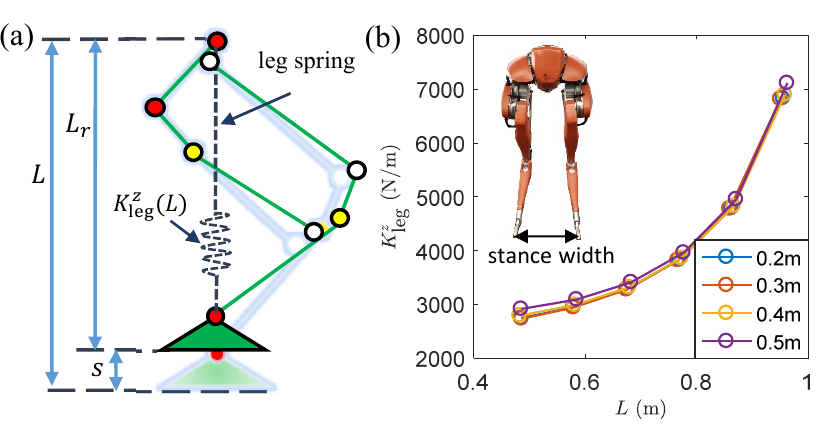}
      \caption{(a) Illustration of the leg spring from the compliant joints. (b) The vertical leg stiffness v.s virtual leg length for different stance width.}
      \label{CassieKine}
\end{figure}

This naturally inspires a definition of \textbf{virtual leg length} $L(q_{\textrm{knee}})$ to approximate $K_{\textrm{leg}}(q_{\textrm{knee}})$ by $K_{\textrm{leg}}(L)$. The real leg length $L_r(q_{\textrm{knee}}, q_{\textrm{shin}}, q_{\textrm{heel}})$ is defined as the distance between the hip pitch joint and the toe joint, whereas the virtual leg length is the real leg length with zero spring deflections,
\begin{equation}
L(q_{\textrm{knee}}) = L_r(q_{\textrm{knee}}, 0, 0),
\end{equation}
as illustrated in Fig. \ref{CassieKine}(a). Due to Cassie's specific leg design, the compliance mainly appears in the direction of the leg. As we are interested in vertical hopping, the last element in $K_{\textrm{leg}}$, denoted by $K^z_{\textrm{leg}}$, is taken as the stiffness of the leg. Fig. \ref{CassieKine}(b) shows how $K^z_{\textrm{leg}}$ changes with $L$ at different static stance configurations. We apply a polynomial regression\footnotemark  to approximate the function $K^z_{\textrm{leg}}(L)$. The leg damping is approximated in the same way.
\footnotetext[1]{in the form of $K^z_{\textrm{leg}}(L) = \beta_0 + \beta_1 L + \beta_2 L^2 + \beta_4 L^4$}


\subsection{The Actuated Spring-Mass Model}
As the total mass of Cassie is concentrated above its hip, we model the pelvis as the point mass and its compliance as the spring attached beneath the mass. As the leg length can change with different kinematic configurations, we model the leg as a massless rod between the point mass and the spring. The spring stiffness and damping are obtained from the abovementioned polynomial regressions. As the leg length can change under motor joint actuation, we define the second order derivative of the leg length $\ddot{L}$ as the virtual actuation. Fig. \ref{SLIP}(a) illustrates the actuated spring-mass model on the ground, where $x=L_r$ is the vertical position of the mass. The dynamics are,
\begin{eqnarray}
&&\ddot{x} = \frac{F^s}{m} - g,\nonumber \\
&&\ddot{s}  = \ddot{L}- \ddot{x} , \nonumber
\end{eqnarray}
 where $s$ is the spring deformation, and $F^s = K(L)s + D(L)\dot{s} $ is the spring force. $K(L)$ and $D(L)$ are the stiffness and damping of the spring, respectively. The kinematic constraints are $s + x = L$ and
$\dot{s} + \dot{x} = \dot{L}$.
\subsection{Jumping and Landing Optimization}
This actuated spring-mass model naturally enables the traditional trajectory optimization framework to function for planning dynamic tasks such as jumping to a desired height and landing to a static configuration. One can parameterize a time-based actuation profile for $\ddot{L}$ and then find the optimal trajectories for the task while minimizing a cost such as the consumed energy. Here, we apply direct collocation methods \cite{hereid20163d} to formulate the trajectory optimization problems; the approach is similar to \cite{hubicki2016tractable}. An even nodal spacing is used for discretizing the trajectory in time. The defect constraint is introduced algebraically by an implicit trapezoidal integration scheme. To enable hopping, we optimize two tasks on the spring-mass model as follows.

\textbf{Jumping.} First, we optimize the spring-mass system to jump to a desired height $x_{\textrm{des}}$ from standing at rest on the ground. When the spring-mass is off the ground, it is in ballistic phase. Therefore, we only discretize the trajectories to the end of ground contact. The task of jumping to the desired height is defined as the equality constraint,
\begin{equation}
x_f + \frac{\dot{x}_f^2}{2g} = x_{\textrm{des}},
\end{equation}
where $x_f$ is the last state of $x$ at standing. Initial states also need to satisfy the initial condition of the system, and the leg length $L$ is constrained to be in the range of kinematic capability of Cassie. Additionally, the ground reaction force, which equals the spring force, must be nonnegative during standing and reaches to zero at the end of ground contact. It is desirable to minimize the virtually consumed energy for this task by defining the cost\footnotemark,
\begin{equation}
J_{\textrm{Jumping}} =\int_{0}^{T} \ddot{L}(t)^2  dt ,
\end{equation}
where $T$ is the duration of the ground contact phase.
\footnotetext{The cost is calculated via trapezoidal integration over time discretization. To simplify the notation, we use $\int$ instead of $\sum$ to avoid introducing additional variables for each discretization.}

\textbf{Landing.} After the ballistic phase, the spring-mass system will land on the ground. One could optimize the trajectory to enable continuous jumping. In our case, we want the system to come to a desired resting configuration, which is enforced by equality constraints on the final states:
\begin{eqnarray}
x_f = x_{\textrm{des}},  \nonumber \\
\dot{L}_f = \dot{s}_f = 0,  \  \ddot{L}_f = \ddot{s}_f =0,  \nonumber\\
 F^s_f - mg = 0, \ddot{L}_f = 0. \nonumber
\end{eqnarray}
Again, the spring force has to be nonnegative. More importantly, we enforce that the spring force must be larger than a desired constant value to increase the feasibility of the trajectory on the full dynamics of the robot, i.e., the robot does not leave off the ground again. Also, the spring deflection $s$ has to be smaller than a certain value so that the spring deflections on the robot are within the ranges of the hardware limits. The cost function can be the same as that in jumping to minimize the consumed energy. We add an extra term to minimize the spring oscillation simultaneously, yielding:
\begin{equation}
J_{\textrm{Landing}} =\int_{0}^{T} \ddot{L}(t)^2 + \alpha \dot{s}(t)^2  dt,
\end{equation}
where $\alpha$ is a weighting coefficient.

\begin{figure}[t]
      \centering
      \includegraphics[width= 3in]{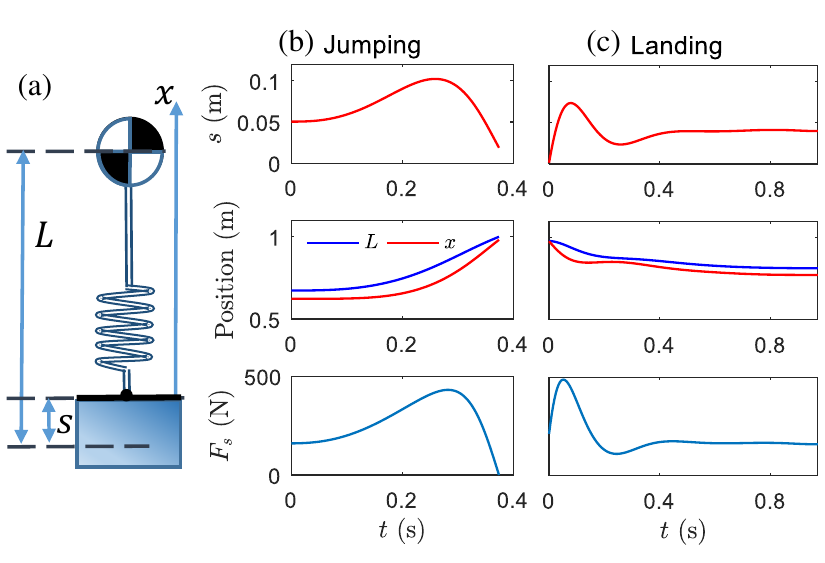}
      \caption{(a) The spring-mass model. (b,c) Results of the jumping and landing trajectory optimization.}
      \label{SLIP}
\end{figure}
As a direct result of the low dimensionality and mild nonlinearity of the system, the optimization is solved in a fast and reliable fashion, typically within 2 seconds. Examples of the optimization results are shown in Fig. \ref{SLIP} (b) and (c).
\section{Controller Synthesis for Hopping}
The hopping of the spring-mass system offers important insights into the intrinsic hopping dynamics of the full-order dynamics from which it was derived.
In this section, we explain how the trajectory of leg length can be encoded via a control Lyapunov Function based quadratic program (CLF-QP). This yields a nonlinear controller that achieves hopping on the biped Cassie---thus a \textit{priori} nonlinear optimization on the full robot \cite{hereid20163d} is not required.
\subsection{The Multi-domain Hybrid System of Hopping}
\begin{figure}[t]
      \centering
      \includegraphics[width= 3.3in]{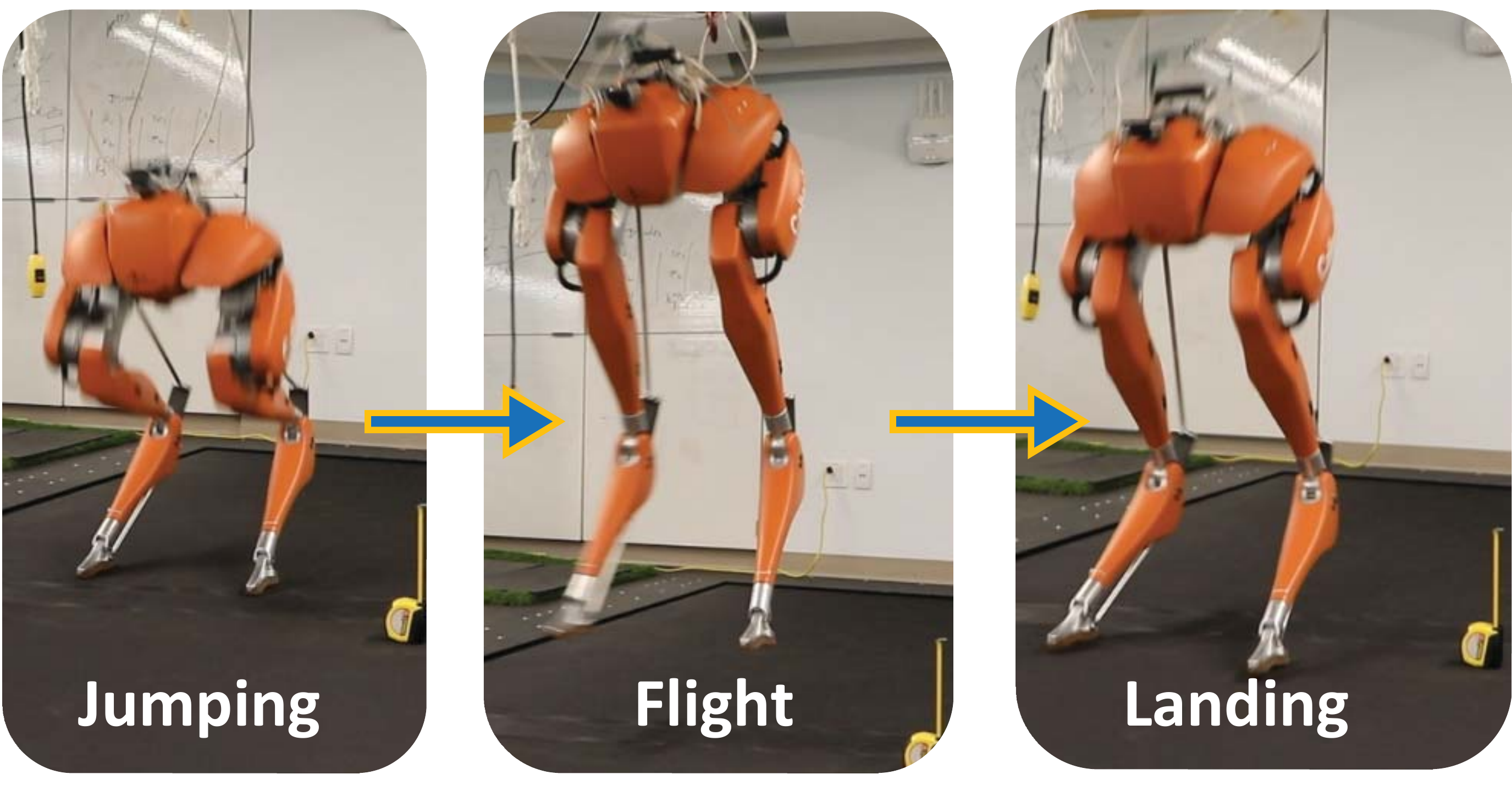}
      \caption{Discrete domains associated with hopping.}
      \label{domains}
\end{figure}

The hopping motion inherently is a hybrid dynamical phenomenon with a ground and flight phase. The behavior will, therefore, be described by a hybrid control system of the form:
\begin{equation}
\mathscr{HC} = (\Gamma,\mathcal{D}, \mathcal{U}, \mathcal{S},\Delta,FG).
\end{equation}
Detailed definitions can be found in \cite{ames2014human}. We assume that the robot only hops in the sagittal plane and its left and right toe always have the same contact mode. It is also desirable to assume that the front and back part of the toe have the same contact modes for simplification purposes. As illustrated in Fig. \ref{domains}, three domains are defined for hopping, i.e. $ \mathcal{D} = \{ \mathcal{D}_J, \mathcal{D}_F, \mathcal{D}_L\}$, where $\{J,F,L\}$ represent \textit{Jumping}, \textit{Flight} and \textit{Landing} respectively. Consequently the directed graph $\Gamma = (V,E)$ is defined by the vertices $V = \{J, F, L \}$ and the edges $E = \{J \rightarrow F,F \rightarrow L \}$.

In \textit{Jumping}, the feet are in contact with the ground. The number of holonomic constraints is $n_{h, J} = 12$. The transition from \textit{Jumping} to \textit{Flight} happens when the ground reaction normal forces cross zero. Thus the domain and associated guard can be defined by,
\begin{eqnarray}
\mathcal{D}_J := \{(q,\dot{q},u): h_{J}(q)= 0, F_z^{\textrm{Foot}}(q,\dot{q},u) >0 \},\\
\mathcal{S}_{J\rightarrow F} :=  \{(q,\dot{q},u): h_{J}(q)= 0, F_z^{\textrm{Foot}}(q,\dot{q},u) =0 \}.
\end{eqnarray}
As there is no impact at the transition from \textit{Jumping} to \textit{Flight}, the reset map is an identity map. 

In \textit{Flight}, the feet are off the ground, $n_{h, F} = 2$, and the transition from \textit{Flight} to \textit{Landing} happens when the feet strike the ground. Therefore, we define the domain and corresponding guard by,
\begin{eqnarray}
\mathcal{D}_F := \{(q,\dot{q},u): \textrm{P}_{z}^{\textrm{Foot}}(q)> 0, F^{\textrm{Foot}}(q,\dot{q},u) = 0 \},\\
\mathcal{S}_{F\rightarrow L} :=  \{(q,\dot{q},u): \textrm{P}_{z}^{\textrm{Foot}}(q)= 0, v_z^{\textrm{Foot}}(q,\dot{q}) <0 \}.
\end{eqnarray}
We model the impact between the feet and the ground as plastic impact, the reset map of which is detailed in \cite{ames2014human}.

In \textit{Landing}, the feet are in contact with the ground again, $n_{h, L} = 12$, and the domain $\mathcal{D}_L$ can be defined as the same as $\mathcal{D}_J$. As we only focus on a single hopping behavior, the system stays in \textit{Landing} after it is reached. There is no need to define its guard.

The continuous dynamics of the system for each domain can be obtained from \eqref{eom} and \eqref{hol}, wherein the exact forms are specified from the corresponding holonomic constraints. Let $h_{\textrm{rod}}, h_{\textrm{Foot}}$ denote the holonomic constraints on the closed kinematic chains and the foot contacts, respectively. Then  $h_J= h_L = \{h_{\textrm{rod}}, h_{\textrm{Foot}}\}$ and $h_F = \{h_{\textrm{rod}}\}$.
\subsection{CLF-QP}
To control the motion of hopping, we suppose that the desired motion is defined by some desired outputs $y^d$ for each domain $v\in V$. For robotic systems, the desired outputs can be outputs with relative degree 1 ($\mathcal{RD}_1$) and degree 2 ($\mathcal{RD}_2$) \cite{khalil1996noninear}. Let $\mathcal{RD}_1$ outputs be represented by $y_1\in \mathbb{R}^{o_1}$ and $\mathcal{RD}_2$ outputs represented by $y_2\in \mathbb{R}^{o_2}$. We assume that the desired motion is of time-based trajectories, thus the outputs can be defined as follows \cite{ames2014human} \cite{khalil1996noninear}:
 \begin{eqnarray}
 &&y_{1}(q,\dot{q},t) = \dot{y}^a_{1}(q,\dot{q}) - y^d_{1}(t),\\
 &&y_{2}(q,t) = y^a_{2}(q) - y^d_{2}(t),
\end{eqnarray}
where the superscript $a$ denotes the actual and $d$ denotes the desired. The objective of the control is to drive $y_1 \rightarrow 0$ and $y_2 \rightarrow 0$.  Differentiating $y_1$ once and $y_2$ twice yields the affine control system on the output dynamics:
\begin{equation}
\begin{bmatrix}
 \dot{y}_1 \\
\ddot{y}_2
\end{bmatrix} =
\underset{\mathbf{\mathcal{L}_f} }{\underbrace{
\begin{bmatrix}
\mathcal{L}_fy_1(q,\dot{q})  - \dot{y}_{1}^d\\
\mathcal{L}^2_fy_2(q,\dot{q}) - \ddot{y}_{2}^d
\end{bmatrix}}}
+
\underset{\mathcal{A}}{\underbrace{
\begin{bmatrix}
\mathcal{L}_gy_1(q,\dot{q})\\
\mathcal{L}_g\mathcal{L}_fy_2(q,\dot{q})
\end{bmatrix}}} u,
\end{equation}
where $\mathcal{A}$ is the decoupling matrix, $\mathcal{L}$ denotes the Lie derivative and $u\in \mathbb{R}^m$ is the control input. The dependency on $t$ is dropped from here to simplify the notation. In case when $u$ can be found to satisfy the following equality:
\begin{equation}
\mathcal{A} u = - \mathbf{\mathcal{L}_f} + \mu,
\label{muAndu}
\end{equation}
the output dynamics becomes this linear control system:
\begin{equation}
\dot{\eta } = \underset{F}{\underbrace{\begin{bmatrix}
0 & 0 & 0\\
0 & 0 & I_2\\
0 & 0  & 0
\end{bmatrix}}} \eta + \underset{G}{\underbrace{\begin{bmatrix}
I_1 & 0\\
0 & 0\\
0 & I_2
\end{bmatrix}}}\mu,
\end{equation}
where $\eta = \left [ y_1, y_2, \dot{y}_2 \right ]^T$, $I_1$ and $I_2$ are identity matrices with dimension $o_1$ and $o_2$ respectively, and $\mu$ is the auxiliary control input \cite{khalil1996noninear}. One can choose $\mu$ to exponentially stabilize the linear system. For example, choosing \cite{ames2013towards}
\begin{equation}
\mu = \begin{bmatrix}
-\epsilon y_1 \\
-2 \epsilon \dot{y}_2 - {\epsilon}^2y_2
\end{bmatrix}
\end{equation}
results the linear output dynamics:
\begin{equation}
\begin{bmatrix} \dot{y_1} \\ \dot{y_2} \\ \ddot{y_2} \end{bmatrix} =
\begin{bmatrix} -\epsilon I_1 & 0 & 0 \\ 0 & 0 & I_2 \\  0 & -\epsilon^2I_2 & -2\epsilon I_2 \end{bmatrix}
\begin{bmatrix} y_1 \\ y_2 \\ \dot{y_2} \end{bmatrix},
\end{equation}
which is exponentially stable when $\epsilon>0$. However, such $\mu$ does not utilize the natural dynamics of the system and oftentimes may not be realizable on the robotic system if there are stringent physical constraints (e.g. torque bounds) that must be enforced.

The above construction motivates constructing rapidly exponentially stabilizing control Lyapunov functions (RES-CLF) \cite{ames2014rapidly} from continuous time algebraic Riccati equations (CARE)\footnotemark or continuous time Lyapunov equations (CTLE)\footnotemark to stabilize the output dynamics exponentially at a chosen rate $\varepsilon$.\footnotetext[3]{$F^T P + PF -PGG^T P + Q = 0.$}\footnotetext[4]{$F^T P + PF + Q = 0.$} Given a solution $P=P^T>0$ to CTLE or CARE with $Q = Q^T>0$, the Lyapunov function is constructed as,
\begin{equation}
V_{\varepsilon}(\eta) = {\eta}^T \underset{ P_{\varepsilon}}{\underbrace{  I_{\varepsilon} P I_{\varepsilon}    }} \eta,
\end{equation}
where $ I_{\varepsilon}  = \textrm{diag}(I_1, \frac{1}{\varepsilon}I_2, I_2)$. The goal of exponential stabilizing $\eta \rightarrow 0$ is encoded by the condition:
\begin{equation}
\dot{V}_{\varepsilon}(\eta) \leq -\frac{\gamma}{\varepsilon} V_{\varepsilon}(\eta),
\label{vdotv}
\end{equation}
with some $\gamma>0$, where,
 \begin{eqnarray}
&& \dot{V}_{\varepsilon}(\eta) = \mathcal{L}_F V_{\varepsilon}(\eta) + \mathcal{L}_G V_{\varepsilon}(\eta) \mu, \label{vdot}\\
&& \ \mathcal{L}_F V_{\varepsilon}(\eta) = {\eta}^T (F^T  P_{\varepsilon} +  P_{\varepsilon} F) \eta,\\
&& \ \mathcal{L}_G V_{\varepsilon}(\eta) = 2{\eta}^T  P_{\varepsilon} G.
\end{eqnarray}
Eq. (\ref{vdotv}) and (\ref{vdot}) indicate an inequality constraint on $\mu$ to achieve exponential stability. This naturally leads to the formulation of quadratic program (QP) to find $\mu$ to minimize the quadratic cost $\mu^T\mu$ . With (\ref{muAndu}), the cost and constraint of the QP can be transformed back onto the original control input $u$ by noting that:
\begin{eqnarray}
\mu^T\mu = u^T \mathcal{A}^T\mathcal{A} u +2\mathbf{\mathcal{L}_f}^T\mathcal{A} u + \mathbf{\mathcal{L}_f}^T \mathbf{\mathcal{L}_f},
\end{eqnarray}
and the inequality from (\ref{vdotv}) and (\ref{vdot}) becomes:
\begin{equation}
 \mathcal{L}_F V_{\varepsilon}(\eta) +\mathcal{L}_G V_{\varepsilon}(\eta)  \mathbf{\mathcal{L}_f}  + \mathcal{L}_G V_{\varepsilon}(\eta) \mathcal{A} u  \leq -\frac{\gamma}{\varepsilon} V_{\varepsilon}(\eta).
\end{equation}
Now the QP can be formulated in terms of solving for $u$ at a current state ($q,\dot{q}$) as follows,
\begin{eqnarray}
 \quad \quad \quad & u^{*} =&\underset{u\in \mathbb{R}^{m}}  {\operatorname{argmin}} \ \  u^T \mathcal{A}^T \mathcal{A}u + 2\mathbf{\mathcal{L}_f}^T\mathcal{A}u \nonumber \\
&\quad \text{s.t.} &A^{\textrm{CLF}}(q,\dot{q}) u \leq b^{\textrm{CLF}}(q,\dot{q}),  \quad   \quad  \quad   \   \quad \text{(CLF)}\nonumber
\end{eqnarray}
where,
\begin{align}
A^{\textrm{CLF}}(q,\dot{q}):=&\mathcal{L}_G V_{\varepsilon}(q,\dot{q}) \mathcal{A}(q,\dot{q}), \\
b^{\textrm{CLF}}(q,\dot{q}):=&-\frac{\gamma}{\varepsilon} V_{\varepsilon}(q,\dot{q})-  \mathcal{L}_F V_{\varepsilon}(q,\dot{q}) \nonumber \\
\quad \quad &- \mathcal{L}_G V_{\varepsilon}(q,\dot{q}) \mathbf{\mathcal{L}_f}(q,\dot{q}).
\end{align}
The result of solving the CLF-QP is a feedback optimal control law to drive the outputs $[\dot{y}^a_{1}(q,\dot{q}) ; y^a_{2}(q) ]$ to follow the desired time based trajectories $[y^d_{1}(t); y^d_{2}(t)]$ with exponentially convergence. This formulation also applies when there are only relative degree 2 outputs to be tracked \cite{hereid2014embedding}. For applications of using CLF-QP on robotic systems, torque bounds and additional nontrivial constraints can be included in the QP \cite{ames2013towards}.

\subsection{Output Definition for Hopping }
To apply the CLF-QP formulation on the multi-domain hybrid control system associated with hopping, we define the outputs with reference output trajectories for each domain separately so that the hopping behavior can be enabled. It is important to define two outputs: leg length and centroidal momentum.

\textbf{Leg Length.} The virtual leg length trajectories $L(t)$ from the jumping and landing of the spring-mass model (see Section \ref{SLIPmodel}) are mainly used as the desired virtual leg length $L^{\textrm{des}}(t)$ on the full robot. The springs on the full robot are expected to behave similarly to the spring on the spring-mass model when the virtual leg length of the full robot follows $L^{\textrm{des}}(t)$. In other words, the underactuation of the springs is expected to behave accordingly so that the robot can jump off the ground and land on the ground.

\textbf{Centroidal Momentum.}
In \textit{Flight}, the only external force is the gravitational force, thus the robot obeys the conservation of angular momentum about its COM, i.e., the centroidal angular momentum \cite{orin2008centroidal}. To keep the control in \textit{Flight} simple, it is desirable to have small centroidal angular momentum when the robot jumps off the ground. The centrodial momentum of a multi-link robotic system can be expressed as \cite{orin2008centroidal}:
\begin{equation}
{H}_G (q, \dot{q})= A_G(q) \dot{q} ,
\end{equation}
where $H_G\in \mathbb{R}^6$ is the centroidal momentum vector and $A_G(q)\in \mathbb{R}^{6 \times n}$ is the centroidal momentum matrix. As we are mainly concerned with vertical jumping, only the pitch angular momentum is selected:
\begin{equation}
H_{\textrm{Pitch}} (q, \dot{q})= A_{\textrm{Pitch}}(q) \dot{q}.
\end{equation}
Note that the pitch centroidal momentum is a relative degree 1 output. Differentiating it once yields,
\begin{eqnarray}
\dot{H}_{\textrm{Pitch}} = \dot{A}_{\textrm{Pitch}}(q,\dot{q}) \dot{q} +  A_{\textrm{Pitch}}(q) \ddot{q}.
\end{eqnarray}
Now we can specify the outputs for each domain.
\subsubsection{Jumping} When two feet are on the ground, 10 additional holonomic constraints are added to the system. As the robot has 16 degrees of freedom (DoF) except for the tarsus joints and spring joints, we need to apply 6 outputs to guide the motion. As noted above, we select the centroidal pitch momentum as the $\mathcal{RD}_1$ output. Left and right leg length are used as two $\mathcal{RD}_2$ outputs to embed the spring-mass jumping dynamics onto the full robot. It is desired to keep the COM position projected onto the center of the support polygon and to avoid yaw motion of the pelvis, which requires $x_{\textrm{com}}\rightarrow 0, y_{\textrm{com}}\rightarrow 0$ and $\phi_{\textrm{yaw}} \rightarrow 0$. Therefore, we define the outputs for \textit{Jumping} as,
\begin{eqnarray}
&& y_1^{J}(q, \dot{q})= H_{\textrm{Pitch}}(q, \dot{q}) - 0 ,\\
&& y_2^{J}(q, t) = \begin{bmatrix} L_{\textrm{L}}(q)\\ L_{\textrm{R}}(q)\\ x_{\textrm{com}}(q) \\ y_{\textrm{com}}(q) \\ \phi_{\textrm{yaw}}(q) \end{bmatrix}   -
 \begin{bmatrix} L^{\textrm{des}}(t) \\L^{\textrm{des}}(t)\\ 0\\ 0 \\0 \end{bmatrix}.
\end{eqnarray}

One can interpret the outputs as these on the 6 Dof of the pelvis (the mass in the spring-mass model). The specific leg length on left and right leg constraints the height and the roll of the pelvis. The specific horizontal COM positions constraint the forward and lateral positions of the pelvis. The pitch momentum output constraints the pitch of the pelvis.

\subsubsection{Flight} The robot is off the ground and the 6 Dof of the floating base is in underactuation. Ten outputs are needed and thus we specify all the motor positions as the outputs:
\begin{equation}y_2^{F}(q) = q_{m} - q_{m}^{\textrm{des}}.\end{equation}
As the centroidal momentum is controlled to be 0 or small in \textit{Jumping}, the robot is not expected to have large whole-body rotation in \textit{Flight}. We simply let the motor positions at the end of \textit{Jumping} be the desired motor positions for \textit{Flight}. The desired toe motor positions are adjusted to keep the toes parallel to the ground.
 \subsubsection{Landing} The outputs are defined similarly to the outputs in \textit{Jumping}. The left and right leg length are selected as two outputs to embed the spring-mass landing dynamics onto the robot. It is not necessary to keep the pitch momentum zero for all time during landing, so we remove the $\mathcal{RD}_1$ output $H_{\textrm{Pitch}}$. Instead, the pelvis pitch is selected. Thus the outputs are defined as:
  \begin{equation}
 y_2^{L}(q,t) = \begin{bmatrix} L_{\textrm{L}}(q) \\ L_{\textrm{R}}(q)  \\ x_{\textrm{com}}(q)  \\ y_{\textrm{com}}(q)  \\ \phi_{\textrm{pitch}}(q)   \\ \phi_{\textrm{yaw}}(q) \end{bmatrix} -
  \begin{bmatrix}  L^{\textrm{des}}(t) \\ L^{\textrm{des}}(t)  \\ x_{\textrm{com}}^{\textrm{des}}(t) \\ 0  \\ \phi_{\textrm{pitch}}^{\textrm{des}}(t)  \\ 0 \end{bmatrix}.
\end{equation}
The desired pelvis pitch trajectory $\phi_{\textrm{pitch}}^{\textrm{des}}(t)$ and desired COM forward trajectory $x_{\textrm{com}}^{\textrm{des}}(t)$ are designed smoothly from the post-impact positions to 0.
\subsection{Main Control Law}
We apply the CLF-QP based feedback control to the hybrid control system associated with hopping. With an eye towards the constrained optimization formulation \cite{ames2013towards}, we rewrite (\ref{eom}) viewing the torque and force as inputs:
\begin{equation}
 M(q)\ddot{q} + \underset{Y(q,\dot{q})}{\underbrace{ H(q,\dot{q}) - J_s^T(q) \tau_s(q)}} =\underset{\bar{B}_v(q)}{\underbrace{\begin{bmatrix} B  &  J_{h,v}^T(q) \end{bmatrix}}} \underset{\bar{u}_v}{\underbrace{\begin{bmatrix} u \\ F_{h,v} \end{bmatrix}}},
\end{equation}
where $\bar{u}_v\in\mathbb{R}^{m+n_{h,v}}$ is the augmented control input. The affine control system can thus be defined as:
\begin{equation}
\dot{x} = f(x) +g_v(x) \bar{u}_v,
\end{equation}
where,
\begin{equation}
f(x) = \begin{bmatrix} \dot{q} \\ -M^{-1}(q)Y(q,\dot{q})\end{bmatrix}, g_v(x) = \begin{bmatrix} 0 \\ M^{-1}(q)\bar{B}_v(q) \end{bmatrix}.
\end{equation}
The reason of including $F_{h,v}$ as the control input is to easily incorporate the holonomic constraints as equality constraints and ground reaction force constraints as inequality constraints in the quadratic program.

\textit{Holonomic Constraint.} Eq. \eqref{hol} can be rewritten as a function of $\bar{u}_v$,
\begin{align}
&A_{h,v}(q,\dot{q}) \bar{u}_v = b_{h,v}(q,\dot{q}), \\
&A_{h,v}(q,\dot{q}) =J_{h,v}(q) M^{-1}(q) \bar{B}_v(q),\\
&b_{h,v}(q,\dot{q}) =J_{h,v}(q) M^{-1}(q)Y(q,\dot{q}) - \dot{J}_{h,v}(q,\dot{q})\dot{q}.
\end{align}

\textit{Ground Reaction Force.} In \textit{Jumping} and \textit{Landing}, the feet contact the ground. The ground reaction forces have to satisfy the physics constraints such as nonnegative normal forces and non-slipping, formulated by $R F^{\textrm{Foot}} \leq 0$, where $R$ is a constant matrix. The constraint on $\bar{u}_v$ can be written as
\begin{equation}
A^{\textrm{GRF}}_v\bar{u}_v \leq b_v^{\textrm{GRF}},
\end{equation}
where $A^{\textrm{GRF}}_{J/L} \bar{u}_{J/L} = R F^{\textrm{Foot}}$, $A^{\textrm{GRF}}_{F} =0$, $b_v^{\textrm{GRF}} = 0$.

\textit{Torque Constraints.} The motor torque must be within the feasible limits of the robot hardware: $u_{lb} \leq u \leq u_{ub}$.

{\bf Optimization-based controller:}
The resulting QP controller for each domain $v \in V$ is given as follows:
\begin{align}
\bar{u}_v^{*} =\underset{\bar{u}_v\in \mathbb{R}^{m+n_{h,v}}, \delta\in \mathbb{R}} {\text{argmin}}& \bar{u}_v^T \mathcal{A}_v^T \mathcal{A}_v \bar{u}_v + 2(\mathbf{\mathcal{L}}_{\mathbf{f},v})^T\mathcal{A}_v \bar{u}_v  + p\delta^2,\nonumber \\
\text{s.t.} \quad & A_v^{\textrm{CLF}}(q,\dot{q}) \bar{u}_v \leq b_v^{\textrm{CLF}}(q,\dot{q})  + \delta, \ \ \text{(CLF)}\nonumber \\
 \quad &A^{\textrm{GRF}}_v\bar{u}_v \leq b^{\textrm{GRF}}_v,   \quad  \quad  \quad   \quad  \quad \quad \ \text{(GRF)}    \nonumber  \\
 \quad  &u_{lb} \leq u \leq u_{ub},  \nonumber  \quad  \quad  \quad   \quad  \quad  \quad  \text{(Torque)}\\
 \quad  &A_{h,v}(q,\dot{q}) \bar{u}_v = b_{h,v}(q,\dot{q}).     \text{(Holonomic)}\nonumber
\end{align}
To increase the feasibility of the QP, we follow the method in \cite{ames2014rapidly} to relax the CLF constraints by introducing $\delta$ and penalizing the relaxation by adding $p \delta^2$ in the cost, with some large positive constant $p$.

\begin{figure}[!b]
      \centering
      \includegraphics[width= 3.4in]{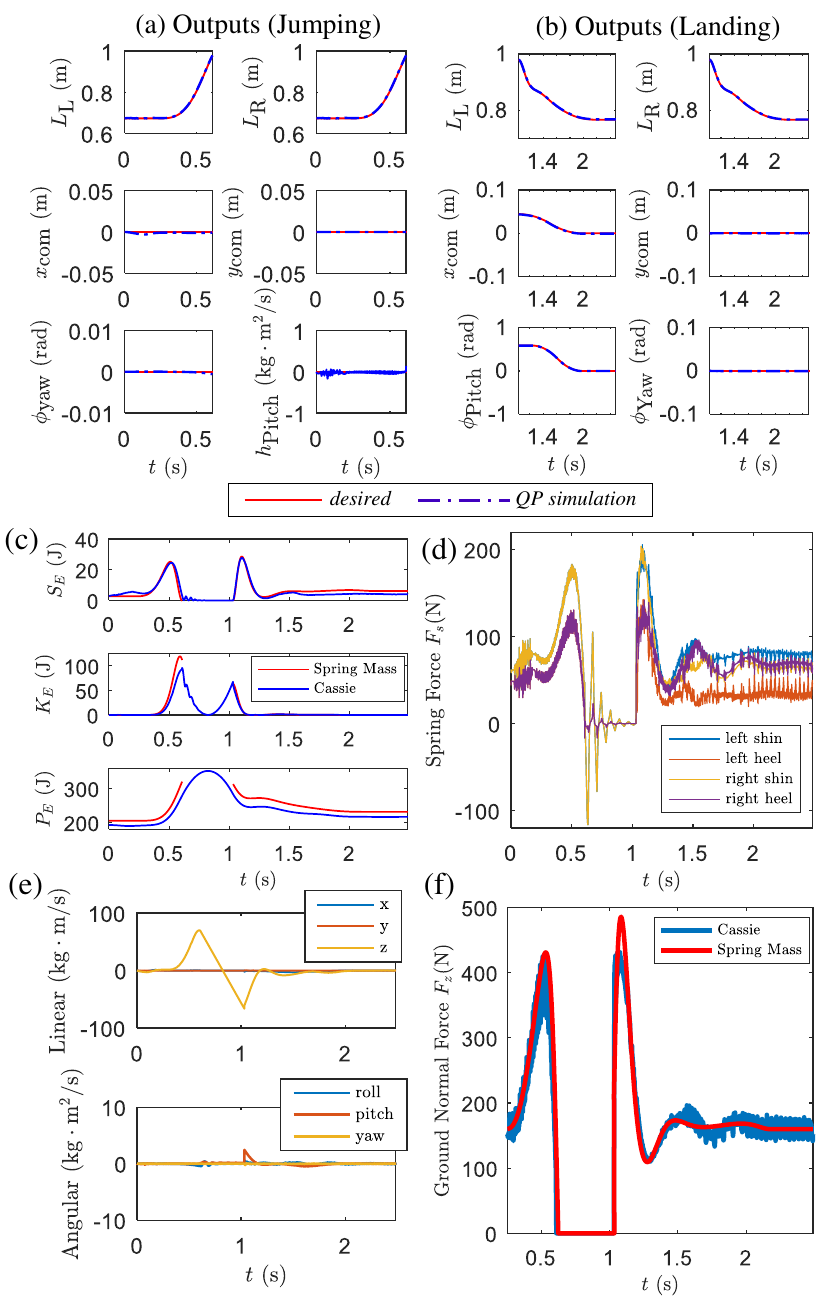}
      \caption{Simulation results. (a,b) Outputs of jumping and landing. (c) Spring energy $S_E$, kinetic energy $K_E$ and gravitational potential energy $P_E$ during hopping. (d) The spring torques of Cassie during hopping. (e) Centroidal momentum of Cassie during hopping. (f) Comparison on the ground reaction normal force of the hopping of Cassie vs that of the spring-mass model. }
      \label{simQP}
\end{figure}
\begin{figure*}[t]
      \centering
      \includegraphics[width = 6.6in]{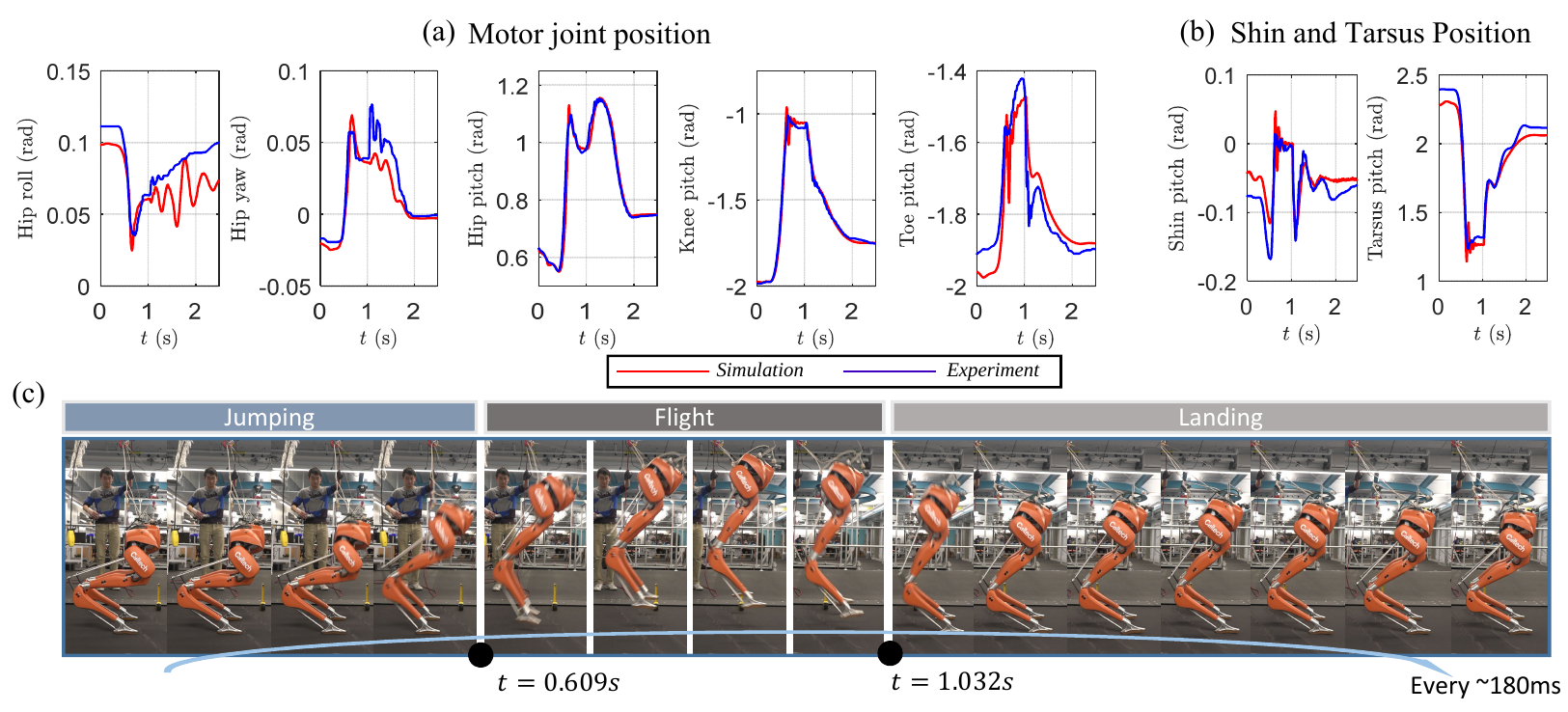}
      \caption{Experiment results. (a) Comparison of motor joint positions on the left leg in simulation vs these in experiment. (b) Comparison of shin and tarsus joints on the left leg in simulation vs these in experiment. (e) Snapshot of the hopping of Cassie.}
      \label{experiment}
\end{figure*}
\section{Simulation and Experimental Results}
To validate our proposed control methodology, we first implement this framework in simulation, and then apply the simulation results on the robot in experiment. The simulation starts from an initial static configuration $q_0$ of the robot. Given a desired apex height, we first optimize the jumping on the spring-mass model from the initial condition corresponding to $q_0$. Then the leg length trajectory is used as the desired output in the CLF-QP. The QP is formulated and solved every at 0.5ms using qpOASES \cite{Ferreau2014}. The dynamics is numerically integrated using MATLAB's ode113 function. At the transition from \textit{Flight} to \textit{Landing}, the post-impact state decide the initial condition for landing optimization on the spring-mass. The leg length trajectory of the landing planning is then used in the CLF-QP for controlling the landing of the robot.

The desired and actual output trajectories are shown in Fig. \ref{simQP} (a) and (b). The simulation indicates fast convergence of output dynamics. Fig. \ref{simQP} (c) shows the comparison on the system energies. Since the robot is not a point mass exactly, the kinetic energy and potential energy differ slightly from these of the spring-mass. Fig. \ref{simQP} (d) and (f) illustrate the spring torques at joints and the ground reaction force (GRF) respectively. Note that the GRF of the robot model align closely with the GRF of the spring-mass model. More importantly, the spring joint forces on the robot show the same profile as the GRF, which equals to the spring force in the spring-mass. This ratifies our hypothesis that the spring of the robot behaves similar to the spring in the spring-mass.

The QP is not yet implemented on the hardware but this is a subject of future work. To validate our method experimentally, we extract the motor joint positions from the simulation and apply position tracking by a PD+feedforward controller on the hardware at 2kHz. The feedforward term is the motor torque from the simulation. Under this setting, the QP is viewed to perform joint trajectory generation. Fig. \ref{experiment} (a) (b) and (c) show the experiment results of a hopping motion with ground clearance of $\sim7$ inches. Due to the model inaccuracy of the physical robot, the spring deflection difference is ineligible and the robot jumps forward consequently. Future work will try to address this issue by utilizing feedback. Experiment videos can be found at \cite{Supplementary}.


\section{Conclusion and Future Work}
This paper presented the planning and controller synthesis to achieve hopping of the bipedal robot Cassie via reduced-order model embedding. The spring-mass model is faithfully established from the mechanical (kinematic and compliant) properties of the robot. The planned trajectories of leg length and additional outputs are rigorously defined in the context of a CLF-QP formulation to facilitate the feedback control of hopping. The realized hopping on Cassie validates the proposed approach.

The future work will be devoted to extending the approach to create different dynamic behaviors on bipedal robots, likely including reduced-order modeling in two or three dimensions for walking and running behaviors. From a theoretical viewpoint, we want to establish stability conditions on how the reduced-order model quantitatively predicts dynamics behaviors of the full-order system. The hope is that this will inform rigorous controller synthesis for complex robots via simple models.

\addtolength{\textheight}{-0.cm}
\bibliographystyle{IEEEtran}
\bibliography{Walking}
\end{document}